%% file: root.tex
\documentclass[letterpaper, 10 pt, conference]{ieeeconf}  

\usepackage[utf8]{inputenc}
\usepackage{textcomp}
\usepackage{graphicx}
\usepackage[version=4]{mhchem}
\usepackage{siunitx}
\usepackage{longtable,tabularx}
\usepackage{accents}
\usepackage{comment}
\usepackage{mathtools}
\usepackage{subfigure}
\usepackage{caption}
\usepackage{color}
\usepackage{amsmath,amssymb}
\usepackage{booktabs}
\usepackage{tabu}
\usepackage{algorithm}
\usepackage{algpseudocode}
\usepackage{hyperref}
\usepackage{soul}
\usepackage[dvipsnames]{xcolor}
\usepackage{multirow}
\usepackage{enumerate}
\usepackage{svg}
\usepackage{epstopdf}
\usepackage{soul}
\usepackage{arydshln}
\usepackage[colorinlistoftodos]{todonotes}
\usepackage{colortbl}

\usepackage{tikz}
\usepackage{siunitx} 
\usetikzlibrary{arrows, arrows.meta}
\usetikzlibrary{graphs}
\usetikzlibrary{snakes,backgrounds,patterns,matrix,shapes,fit,calc,shadows,plotmarks} 
\usetikzlibrary{bayesnet}

\usepackage{standalone}

\tikzstyle{target}=[draw,fill=yellow!50,circle,minimum size=16pt,inner sep=0pt]

\tikzstyle{output}=[draw,fill=blue!50,circle,minimum size=16pt,inner sep=0pt]
\tikzstyle{bias}=[draw,fill=gray!50,circle,minimum size=20pt,inner sep=2pt]
\tikzstyle{arrow}=[arrows={{Latex[scale=0.5]}-}, thick]  
\tikzstyle{box}=[rectangle, draw=black!100] 

\tikzset{
    between/.style args={#1 and #2}{
         at = ($(#1)!0.5!(#2)$)
    }
}

\IEEEoverridecommandlockouts                              

\title{\LARGE \bf
Nonlinear Heterogeneous Bayesian Decentralized Data Fusion
}

\author{Ofer Dagan, Tycho L. Cinquini and Nisar R. Ahmed
\thanks{*This work was partially supported by NASA STTR award 80NSSC20C0314. 
}
\thanks{
The authors are with the Smead Aerospace Engineering Sciences Department, University of Colorado Boulder, Boulder, CO 80309 USA {\tt\small ofer.dagan@colorado.edu; Nisar.Ahmed@colorado.edu}}
}

\begin{document}

\maketitle
\thispagestyle{empty}
\pagestyle{empty}

\begin{abstract}
The factor graph decentralized data fusion (FG-DDF) framework was developed for the analysis and exploitation of conditional independence in \emph{heterogeneous} Bayesian decentralized fusion problems, in which robots update and fuse pdfs over different, but overlapping subsets of random states. 
This allows robots to efficiently use smaller probabilistic models and sparse message passing to accurately and scalably fuse relevant local parts of a larger global joint state pdf while accounting for data dependencies between robots. 
Whereas prior work required limiting assumptions about network connectivity and model linearity, this paper relaxes these to explore the applicability and robustness of FG-DDF in more general settings. 
We develop a new heterogeneous fusion rule which generalizes the homogeneous covariance intersection algorithm for such cases and test it in multi-robot tracking and localization scenarios with non-linear motion/observation models under communication dropouts. 
Simulation and hardware experiments show that, in practice, the FG-DDF continues to provide consistent filtered estimates under these more practical operating conditions, while reducing computation and communication costs by more than $99\%$, thus enabling the design of scalable real-world multi-robot systems.  

\end{abstract}

\section{INTRODUCTION}
\input{Text/0_Introduction.tex}

\section{Problem Statement}
\label{sec:probStatement}
\input{Text/2_ProblemStatement}

\section{Technical Approach}
\label{sec:techApproach}
\input{Text/3_Technical_approach}

\section{Experiments}
\label{sec:empStudy}

\input{Text/4_SimulationStudy}

\label{sec:sim}

\subsection{Hardware Experiment}
\label{sec:exp}
\input{Text/5_HardwareExperiments}

\section{Conclusions}
\label{sec:conclusions}
\input{Text/6_Conclusions}
\bibliographystyle{IEEEtran}
\bibliography{references.bib}

\end{document}

%% file: Text/0_Introduction.tex
Bayesian decentralized data fusion (DDF) \cite{chong_distributed_1983} is applicable to networks of robots acting in a shared (problem) space toward common goals that require estimation over a global set of random variables (rvs). 
Robots can gain new data from local sensors and by peer-to-peer communication of their current local estimated joint probability distribution function (pdf), often described only by their mean and covariance over the full, \emph{homogeneous}, set of rvs.
\emph{Heterogeneous} DDF is the sub-class of DDF problems where communicating robots fuse information with respect to pdfs over two non-equal, but overlapping, subsets of the rvs \cite{dagan_exact_2023}. 

Many collaborative applications across robotics are instances of heterogeneous fusion. Heterogeneous fusion enables the scalable operation of large robotic teams by distributing the global joint inference problem to smaller, overlapping, local ones. 
Thus robots are able to reason over their local inference task and communicate only relevant data to their neighbors.
For example: (i) in multi-robot simultaneous localization and mapping (SLAM) \cite{cunningham_ddf-sam_2013}, robots keep their estimated positions local, and share only parts of the map; (ii) in multi-robot tracking with sensor bias uncertainties, robots can share estimates over common targets, while non-mutual targets and sensor biases are only estimated locally \cite{dagan_exact_2023}; (iii) when estimating local sensor measurement bias and the temperature distribution across a room, then bias estimates are kept local while temperature estimates are shared \cite{paskin_robust_2004}. 
Thus, enabling each robot to reason over and communicate only parts of the full joint inference problem is imperative for scalability, 
as local communication and computation requirements now scale with each robot's local inference task and not with the full global inference problem.
This `divide and conquer' approach can lead to more than $95\%$ reduction in communication and computation costs  \cite{dagan_exact_2023},\cite{dagan_factor_2021}.



One of the main challenges in DDF is to correctly account for dependencies in the data gathered and shared by the robots so that new data is treated as new only once. 
In classical homogeneous DDF, where robots communicate and infer the same global set of rvs, methods exist to account for such dependencies either implicitly (e.g., covariance intersection (CI) \cite{julier_non-divergent_1997}, inverse covariance intersection (ICI) \cite{noack_decentralized_2017}) or explicitly (e.g., by using a channel filter (CF) \cite{grime_data_1994}).
However, some fundamental issues arise for nonlinear systems -- namely, the definition and meaning of common data dependencies become less clear, e.g., when robots propagate their pdfs using linearization based on different state estimates. 

In heterogeneous DDF the problem becomes more acute as:
(i) robots' state vectors are non-equal, thus their linearization points are inherently different; 
(ii) in addition to homogeneous \emph{known unknown} dependencies, there are now (hidden) \emph{unknown unknown} dependencies between non-mutual rvs that must be treated \cite{dagan_conservative_2022}; 
(iii) it is neither obvious nor clear what impact real-world issues such as imperfect communication (message dropouts) have on heterogeneous DDF since these can lead to robots having different perspectives on common data dependencies.


In this paper, we build on and extend our previous work by (i) proposing a new CI fusion rule (HS-CI) for heterogeneous DDF \cite{dagan_exact_2023};  
(ii) exploring the robustness and applicability of the factor graph-based fusion framework (FG-DDF) \cite{dagan_factor_2021} in simulations and hardware experiments, running onboard Clearpath Jackal unmanned ground vehicle (UGV), and under realistic, non-linear, heterogeneous multi-robot scenarios. 
(iii) demonstrating that although it is not theoretically clear how common data dependencies are exactly tracked in non-linear systems, in practice, the FG-DDF framework yields consistent estimates at each robot, even when $50\%$ of the messages do not get to their destination. 
These contributions enable the design of scalable real-world heterogeneous multi-robot collaborative applications. 

The rest of the paper is organized as follows. Sec. \ref{sec:probStatement} defines the heterogeneous fusion problem and reviews related work. Sec. \ref{sec:techApproach} presents the technical approach for nonlinear heterogeneous DDF and develops a new heterogeneous CI rule. 
Sec. \ref{sec:empStudy} details the empirical multi-robot simulation and hardware studies. Sec. \ref{sec:conclusions} summarizes the findings and describes future work.

%% file: Text/2_ProblemStatement.tex
Consider a network of $\left| N_r\right|=n_r$ autonomous robots, jointly monitoring a global set of (possibly time-dependent) rvs $\chi_k$, at time step $k$.
Each robot $i\in N_r$ is tasked with inferring an overlapping subset of rvs $\chi^i_k\subset \chi_k$ and uses Bayes' rule to recursively update its local prior pdf over $\chi^i_k$.
We can split these updates into two modules: 
(1) a filter module, which includes the familiar prediction--marginalization--measurement update; and (2) a fusion module, which includes fusing data from neighboring robots.
In every time step, each robot can thus update its pdf using the following steps: (i) prediction, using the conditional transition probability $p(\chi^i_k|\chi^i_{k-1})$, (ii) marginalization over rvs from the previous time step; (iii) Bayesian fusion of local sensor data $y^i_k$, described by the conditional likelihood $p(y^i_k|\chi^i_k)$; and (iv) fusion with any neighboring robot $j\in N^i_r$ by exchanging pdfs over their common rvs via the peer-to-peer \emph{Heterogeneous State} (HS) fusion rule \cite{dagan_exact_2023}, 
\begin{equation}
    \begin{split}
        p_f^i(\chi^i|&Z^{i,+}_k)\propto \\
&\underbrace{\frac{p^i(\chi^{ij}_C|Z^{i,-}_k)p^j(\chi^{ij}_C|Z^{j,-}_k)}{p^{ij}_c(\chi^{ij}_C|Z^{i,-}_k \cap Z^{j,-}_k)}}_{p_f(\chi^{ij}_C|Z^{i,+}_k)}  
         \cdot p^i(\chi^{i\backslash j}|\chi^{ij}_C,Z^{i,-}_k),
    \end{split}
    \label{eq:Heterogeneous_fusion}
\end{equation}
where $p_f^i(\cdot)$ is the fused posterior pdf at robot $i$.
$p^{ij}_c(\chi^{ij}_C|Z^{i,-}_k \cap Z^{j,-}_k)$ is the pdf over robots $i$ and $j$ common rvs, given their common data, which can stem from common prior, dynamic models, and previous communication episodes. The rest of the notation is defined in Table \ref{tab:key_notations}. Note that when $\chi^i=\chi_C^{ij}=\chi$, (\ref{eq:Heterogeneous_fusion}) degenerates to the homogeneous Bayesian fusion rule \cite{chong_distributed_1983}.

\begin{table}[b]
\vspace{-0.2in}
\renewcommand{\arraystretch}{1.5}
\caption{Key notations and definitions used in the paper }
    \begin{center}
    \begin{tabular}{l l}
       \textbf{ Variable Name }& \textbf{Description} \\ \hline 
        $N_r$ & Set of $n_r$ robots \\ \arrayrulecolor[gray]{0.8}\hline
        $N_r^i$ & Set of robot $i$'s (network) neighbors \\ \hline
        $\chi^i$ & Robot $i$'s set of rvs \\ \hline
        $\chi^i_L$ & Local rvs, only monitored by $i$ \\ \hline
        $\chi^{ij}_C = \chi^i\cap\chi^j$ & Set of rvs common to $i$ and $j$ \\ \hline
        $\chi^i_C=\bigcup_{j\in N_r^i}^{}\chi^{ij}_C$ & Common rvs to $i$ with any neighbor  \\ \hline
        $\chi^{i\setminus j}=\chi^i_L\cup\{\chi^i_C\setminus \chi^{ij}_C \}$ & Non-mutual rvs to $i$ and $j$ \\ \hline
        $Z^{i,-}_k$ & $i$'s available data at time $k$ prior to fusion \\ \hline 
        $Z^{i,+}_k$ & $i$'s available data at time $k$ post fusion \\ \hline
    \end{tabular}
    \end{center}
    \label{tab:key_notations}
    \vspace{-0.1in}
\end{table} 

There are two key points pertaining to the fusion rule in (\ref{eq:Heterogeneous_fusion}). First, non-mutual variables have to be conditionally independent given common variables between the communicating robots, i.e., $\chi^{i\setminus j}\perp \chi^{j\setminus i}|\chi^{ij}_C$, for it to be valid \cite{dagan_exact_2023}. 
The second point -- which lies at the core of DDF problems in general and heterogeneous DDF specifically -- is how to account for dependencies in the data shared between robots so that new data are treated as such only once. 
In heterogeneous DDF, these dependencies are accounted for in the `common' pdf $p^{ij}_c(\chi^{ij}_C|Z^{i,-}_k \cap Z^{j,-}_k)$, in the denominator of (\ref{eq:Heterogeneous_fusion}). 
In the DDF literature, this `common' pdf is treated either with exact methods, which explicitly track dependencies in the data, e.g., by adding a channel filter (CF) \cite{grime_data_1994}, or approximate methods such as covariance intersection (CI) \cite{julier_non-divergent_1997} and inverse covariance intersection (ICI) \cite{noack_decentralized_2017}, where unknown dependencies between the estimates are removed at the cost of inflating the covariance matrix.      

To address these challenges, we develop a factor graph-based framework, dubbed FG-DDF \cite{dagan_factor_2021}. 
The FG-DDF framework enables theoretical analysis and exploitation of the conditional independence structure in robotic heterogeneous fusion problems. It can be used in practice as the robot's inference engine, and as a tool to explicitly track dependencies in the data.  
For example, it is used in \cite{dagan_conservative_2022} to develop a method for recursive conservative filtering for heterogeneous fusion in dynamic systems. 

Since the aforementioned body of work aimed at gaining a fundamental understanding of the heterogeneous DDF problem and the nature of dependencies in the data held by the robots in the network, several assumptions were made: 
\begin{enumerate}
    \item The dynamic system transition and observation models, $p(\chi^i_k|\chi^i_{k-1})$ and $p(y^i_k|\chi^i_k)$, respectively, are \emph{linear} with additive white Gaussian noise (AWGN).
    \item The network communication topology is described by an undirected acyclic graph. 
    \item Perfect communication, that is, every message sent by a robot $i$ is received by the neighboring robot $j$.
\end{enumerate}
\noindent
These allow for the pdfs over common variables between any two robots $p^{ij}_c(\chi^{ij}_C|Z^{i,-}_k \cap Z^{j,-}_k)$ to be explicitly tracked and removed in fusion (\ref{eq:Heterogeneous_fusion}) using the heterogeneous-state channel filter (HS-CF) \cite{dagan_exact_2023}. 
It has been shown, that under these assumptions, the FG-DDF framework enables robots to accurately process (infer) and communicate only parts of the global joint pdf. 
Which, results in a communication and computation reduction of more than $95\%$ in linear-Gaussian problems \cite{dagan_exact_2023},\cite{dagan_factor_2021}, and lets heterogeneous robotic teams scale much more effectively.

However, questions arise as to how robust/applicable the FG-DDF framework developed in \cite{dagan_factor_2021} and \cite{dagan_conservative_2022} is to: \emph{non-linear} transition and observation models; real-world problems such as message dropouts; and approximations of the `common' pdf, $p^{ij}_c(\chi^{ij}_C|Z^{i,-}_k \cap Z^{j,-}_k)$, e.g., via a heterogeneous version of the covariance intersection (CI) algorithm. This paper explores those questions using simulations and hardware experiments.

\textbf{Related work:}
In heterogeneous fusion problems, marginalization often couples previously conditionally independent rvs.    
Since maintaining conditional independence is key to the solution approach, we identified two main aspects of the problem that affect the solution:  
(i) type of common ($\chi^i_C$) and local variables ($\chi^i_L$), i.e., whether they are dynamic or static; (ii) whether the inference algorithm solves for static variables, dynamic variables with a smoothing approach, or dynamic variables with a filtering (recursive) approach.  

In \cite{paskin_robust_2004}, Paskin and Guestrin describe a distributed junction tree (D-JT) algorithm to infer the temperature field in a lab setting, in the presence of local sensor measurement bias. 
Here all variables are static, and sensors estimate and share a subset of static temperature variables while keeping their bias estimates local. 
Thus, the conditional independence structure, in this case, is not affected by marginalization and stays constant in time. 
Makarenko \textit{et al.} \cite{makarenko_decentralised_2009} extend the D-JT to include dynamics and cast it as a DDF problem. 
But their algorithm is only applied to a single variable of interest, i.e., it describes a homogeneous fusion problem, which does not necessitate maintaining a conditional independence structure. 
In \cite{cunningham_ddf-sam_2013}, Cunningham \textit{et al.} develop a smoothing and mapping (SAM) technique called DDF-SAM, based on factor graphs. 
As their work focuses on SAM, the shared variables are static (a subset of the map) and the solution (which includes the dynamic robot states) is smoother-based, i.e., the algorithm circumvents the challenges resulting from marginalizing over past states, as the map variables are independent given the full robot trajectory. 
Chong and Mori \cite{chong_graphical_2004} use information graphs and Bayes nets to analyze and design algorithms for nonlinear distributed estimation. 
Their work presents a wide analysis of the problem to maintain conditional independence in the dynamic case, but assumes a deterministic state process, and does not account for stochastic dynamic problems.

From the heterogeneous fusion perspective, a filtering solution to a stochastic dynamic system (with dynamic local and common variables) is a more general and challenging scenario as it becomes harder to: (i) correctly remove common data in fusion, as it is `rolled up' into the current estimate upon marginalization over past states (especially when it was propagated through a non-linear transformation), and (ii) maintain conditional independence between non-mutual states.  
For these reasons, this paper uses the combined problems of dynamic multi-target tracking and self-localization as a test case to explore the heterogeneous FG-DDF framework. 
In this scenario, robots independently localize themselves based on range and bearing measurements to known landmarks, while sharing state estimates on common tracked dynamic targets.

%% file: Text/3_Technical_approach.tex

In this section, after a brief introduction to factor graphs in the context of DDF, we provide a summary of the main technical details of the CF and CI methods, and how they are extended for heterogeneous DDF. 
These methods will be used to test the FG-DDF framework in dynamic nonlinear systems, under realistic conditions.

\subsection{Factor Graphs for DDF (FG-DDF)}
A factor graph is a type of probabilistic graphical model (PGM) \cite{frey_factor_1997}, which has gained popularity in the robotics community since it naturally expresses the sparse information dependency structure inherent in many robotic applications \cite{dellaert_factor_2021}. 
A factor graph is an undirected bipartite graph $\mathcal{G}=(F,V,E)$, factorized into smaller functions given by factor nodes $f_l\in F$. 
Each factor $f_l(V_l)$ is connected by edges $e_{lm}\in E$ only to the function's random variables $v_m\in V_l \subset V$.
The joint distribution over the graph is then proportional to the global function $g(V)$,
\begin{equation}
    p(V)\propto g(V)=\prod_{l}f_l(V_l).
    \label{eq:factorization}
\end{equation}

Recent work used factor graphs to exploit conditional independence in heterogeneous DDF problems and suggested a new framework, FG-DDF, to analyze and solve them in static \cite{dagan_factor_2021} and dynamic \cite{dagan_conservative_2022} linear systems. 
In FG-DDF, new factors are added to the graph due to prediction, observation, and fusion, see Fig. \ref{fig:factorgraph_steps}. 
In filtering, the graph is manipulated, or re-factorized, to maintain the conditional independence structure and ensure the estimate is conservative by deflating the factors information matrix (i.e., inflating the covariance) by a deflation constant $\lambda$. For a formal definition and intuition on the deflation constant see \cite{dagan_conservative_2022}.

\input{Figures/factorGraph_steps}

These works assume linear models and Gaussian noise, where each factor is expressed using the information (canonical) form of the Gaussian distribution, i.e., $f_l(V_l)\propto \mathcal{N}^{-1}(V_l; \zeta_l,\Lambda_l)$, where $\zeta_l$ and $\Lambda_l$ are the information vector and matrix, respectively. 
This paper explores the heterogeneous DDF problem in \emph{nonlinear} robotic systems, i.e. its focus is on the fusion module for nonlinear filtering problems. 
In many of these systems, the underlying filter at each robot propagates the first two moments (mean and covariance) of the full pdf through nonlinear transformations, $p(\chi^i_k|\chi^i_{k-1})$ and $p(y^i_k|\chi^i_k)$. 
This can be done using, for example, the extended-Kalman filter (EKF), or the unscented-Kalman filter \cite{julier_unscented_2004}.
We can again use the information form of these filters, e.g., in this paper, we implemented the extended information filter (EIF) \cite{eustice_exactly_2005} equations to add factors into the local robot's graph, this means that the factors are still described by the information vector and matrix.
Note that now the factor graph represents the first two moments of the pdf, and not the full pdf itself.
\subsection{Channel Filter}
In networks with an undirected acyclic communication graph, there is only one communication path between any two robots. 
For such networks, \cite{grime_data_1994} suggests adding a filter, dubbed the channel filter (CF), on the communication channel between every pair of communicating robots, $i$ and $j$, to explicitly calculate $p^{ij}_c(\chi|Z^{i,-}_k \cap Z^{j,-}_k)$ over the full (homogeneous) set of rvs $\chi$. 
In \cite{dagan_exact_2023} we extend this idea to heterogeneous DDF with the HS-CF. In HS-CF, the CF recursively computes the marginal pdf $p^{ij}_c(\chi^{ij}_C|Z^{i,-}_k \cap Z^{j,-}_k)$ in (\ref{eq:Heterogeneous_fusion}), which is then removed from the robot's local marginal pdf. 
In problems where the pdfs are expressed using the marginal information vector ($\bar{\zeta}_{\chi_C^{ij}}$) and matrix ($\bar{\Lambda}_{\chi_C^{ij}\chi_C^{ij}}$), representing the mean and covariance of the pdf, the fused pdf over the subset of common rvs $\chi^{ij}_C$, shown in the left part of (\ref{eq:Heterogeneous_fusion}) is $p_f(\chi^{ij}_C|Z^{i,+}_k)\sim \mathcal{N}^{-1}(\chi^{ij}_C; \bar{\zeta}_{\chi_C^{ij},f}, \bar{\Lambda}_{\chi_C^{ij}\chi_C^{ij},f})$, with   \begin{equation}
    \begin{split}
        &\bar{\zeta}_{\chi_C^{ij},f} = \bar{\zeta}^i_{\chi_C^{ij}}+\bar{\zeta}^j_{\chi_C^{ij}}-\bar{\zeta}_{\chi_C^{ij},c}^{ij},\\ 
        &\bar{\Lambda}_{\chi_C^{ij}\chi_C^{ij},f} = \bar{\Lambda}^i_{\chi_C^{ij}\chi_C^{ij}}+\bar{\Lambda}^j_{\chi_C^{ij}\chi_C^{ij}}-\bar{\Lambda}^{ij}_{\chi_C^{ij}\chi_C^{ij},c}.
        \label{eq:marginalCF}
    \end{split}
\end{equation}
In practice, each robot maintains another factor graph on every communication channel, representing the CF with its neighboring robots. 
In other words, each robot adds and removes factors from its local graph as described in Fig. \ref{fig:factorgraph_steps}(a)-(c), while its CFs use steps (a)-(b). Step (d) in the figure is described by (\ref{eq:marginalCF}) or (\ref{eq:homogeneousCI})-(\ref{eq:marginalCI}) for problems that use the information vector and matrix to describe their pdf.

In many nonlinear problems across robotics, steps (a) and (c) in  Fig. \ref{fig:factorgraph_steps} are done using linearization, e.g., see the EIF \cite{eustice_exactly_2005}. 
In that case, each robot (and their CFs) might use different linearization points to compute the required Jacobians. 
Thus there are no guarantees, even for the homogeneous case, that the pdf over the common data, held by the CFs is propagated exactly the same by both robots. 
In heterogeneous fusion, this is exacerbated, since linearization points are, by definition, different, as the HS-CF only holds data about the marginal pdf over the common state variables for each robot. 
We show that despite the gap in formal guarantees, the HS-CF still provides conservative results for nonlinear heterogeneous systems in our simulated and hardware test cases.


\subsection{Covariance Intersection}
Covariance intersection (CI) \cite{julier_non-divergent_1997} is a widely used approximate method for cyclic or ad-hoc communication topologies. CI computes the weighted average of the robots' information vector and matrix, where the weight, $\omega$, is calculated to optimize some predetermined cost function, e.g., the determinant or trace of the fused covariance matrix. The CI fusion rule is then given by,
\begin{equation}
    \begin{split}
        &\zeta_{f} = \omega\zeta^i+(1-\omega)\zeta^j=\zeta^i+\zeta^j-\overbrace{\big[(1-\omega)\zeta^i+\omega \zeta^j \big ]}^{\zeta^{ij}_c},    \\
        &\Lambda_{f} = \omega\Lambda^i+(1-\omega)\Lambda^j = \Lambda^i+\Lambda^j-\underbrace{\big[(1-\omega)\Lambda^i+\omega \Lambda^j \big ]}_{\Lambda^{ij}_c},
        \label{eq:homogeneousCI}
    \end{split}
\end{equation}
where we show that the information vector and matrix of the `common' pdf $p^{ij}_c(\chi|Z^{i,-}_k \cap Z^{j,-}_k)$ can be approximately evaluated using $\omega$. With this interpretation, we can replace the CF calculated $\bar{\zeta}_{\chi_C^{ij},c}^{ij}$ and $\bar{\Lambda}^{ij}_{\chi_C^{ij}\chi_C^{ij},c}$ in (\ref{eq:marginalCF}) with,
\begin{equation}
    \begin{split}
        &\bar{\zeta}_{\chi_C^{ij},c}^{ij} = (1-\omega)\bar{\zeta}^i_{\chi_C^{ij}}+\omega \bar{\zeta}^j_{\chi_C^{ij}},   \\
        &\bar{\Lambda}^{ij}_{\chi_C^{ij}\chi_C^{ij},c} = (1-\omega)\bar{\Lambda}^i_{\chi_C^{ij}\chi_C^{ij}}+\omega \bar{\Lambda}^j_{\chi_C^{ij}\chi_C^{ij}}.
        \label{eq:marginalCI}
    \end{split}
\end{equation}
With these definitions, we develop a new implicit (approximate)  heterogeneous fusion rule corresponding to (\ref{eq:Heterogeneous_fusion}), the \emph{HS-CI} fusion rule. 
Unlike the original CI fusion rule \cite{julier_non-divergent_1997}, which takes a weighted average of the information vector and matrix of the full homogeneous state vector (including data that is unique to each robot) the above heterogeneous CI fusion rule only `discounts' data over the common states, leaving non-common states (variables) untouched. 
Note that this differs from the split-CI fusion rule \cite{julier_general_2009}, which assumes the data can be separated into dependent and independent parts. This is a much more restrictive assumption than requiring conditional independence, as in the new HS-CI rule.

%% file: Figures/factorGraph_steps.tex
\begin{figure}[tb]
\begin{tikzpicture}[ new set=import nodes]
 \begin{scope}[nodes={set=import nodes}]
            
      \node [text width=4.25cm] at (0.2, 2.5) (block1) {$(a)$ \scriptsize{Prediction --} \scriptsize{Adding factors and variable nodes to the prior pdf. }};
      \node (s1)[latent, minimum size=25pt] at (-1.2,0.75) {$\chi_{L,0}^i$};
      \node [factor, fill=red!100,right=of s1,xshift=0.55cm, label={$f^{1}_{L}(\chi_{L,1}^i|\chi_{L,0}^i)$}] (f1L) {};
      \node (s2) [latent, right=of f1L]  {$\chi_{L,1}^i$};
      \node [factor, above=of s1, label=right:{$f^{0}_L(\chi_{L,0}^{i})$}]  (f0L) {};

      \node [latent, below=of s1, yshift=0.75cm] (x1) {$\chi_{C,0}^{i}$};
      \node [factor, fill=red!100,right=of x1, xshift=0.55cm, label={$f^{1}_{C}(\chi_{C,1}^i|\chi_{C,0}^i)$}] (f1) {};
      \node [latent, right=of f1] (x2) {$\chi_{C,1}^{i}$};
      \node [factor, below=of x1, label=right:{$f^{0}_C(\chi_{C,0}^{i})$}]  (f0) {};

      \node [text width=4.0cm] at (4.55,2.5) (block2) {$(b)$ \scriptsize{Marginalization --} \scriptsize{``Summarizing" factors into a new factor.}};
      \node (s3)[latent, minimum size=25pt] at (4.75,0.75) {$\chi_{L,1}^{i}$};
      \node [factor,fill=red!100, left=of s3, xshift=-0.2cm, label={$f^{2}_L(\chi_{L,1}^{i})$}]  (f2L) {};
      \node (x3)[latent,  below=of s3, yshift=0.75cm] {$\chi_{C,1}^{i}$};
      \node [factor,fill=red!100, left=of x3, xshift=-0.2cm, label={$f^{2}_C(\chi_{C,1}^{i})$}]  (f2) {};

      \node [text width=4.25cm] at (0.2, -2.25) (block3) {$(c)$ \scriptsize{Update --} \scriptsize{Relative measurement factor between local and common rvs. }};
      \node (s4)[latent, minimum size=25pt] at (0,-3.5) {$\chi_{L,1}^{i}$};
      \node [factor, left=of s4, xshift=-0.2cm, label={$f^{2}_L(\cdot)$}]  (f3L) {};
      \node (x4)[latent,  below=of s4, yshift=0.25cm] {$\chi_{C,1}^{i}$};
      \node [factor, left=of x4, xshift=-0.2cm, label={$f^{2}_C(\cdot)$}]  (f3) {};
      \node [factor,fill=red!100, between=s4 and x4, label=right:{$f^{3}(y^i_1|\chi_{L,1}^{i},\chi_{C,1}^{i})$}]  (f4) {};
      
      \node [text width=4.25cm] at (4.55,-2.25) (block4) {$(d)$ \scriptsize{Fusion --} \scriptsize{Factor sent by a neighboring robot over common rvs}};
      \node (s5)[latent, minimum size=25pt] at (4.75,-3.5) {$\chi_{L,1}^{i}$};
      \node [factor, left=of s5, xshift=-0.2cm, label={$f^{2}_L(\cdot)$}]  (f5L) {};
      \node (x5)[latent,  below=of s5, yshift=0.25cm] {$\chi_{C,1}^{i}$};
      \node [factor, left=of x5, xshift=-0.2cm, label={$f^{2}_C(\cdot)$}]  (f5) {};
      \node [factor, between=s5 and x5, label=right:{$f^{3}(\cdot)$}]  (f6) {};
      \node [factor, fill=red!100, below=of x5, label=right:{$f^{4}_C(\chi_{C,1}^{i})$}]  (f6C) {};
     
  \end{scope}
  
 \graph {
    (import nodes);
   
    {x1,x2}--f1, {s1,s2}--f1L,
    s3--f2L, x3--f2,
    s1--f0L, x1--f0,
   
    s4--f3L, x4--f3,
    {s4,x4}--f4,

    {s5,x5}--f6, s5--f5L, x5--f5,
    x5--f6C,
       
    };
    
\end{tikzpicture}
\caption{Four Bayesian update steps for the local and common rv sets, where new factors are shown in red. We choose the explicitly represent dependencies and measurements for clarity in the factor's parentheses \cite{frey_extending_2002}. }
      \label{fig:factorgraph_steps}
      \vspace{-0.2in}
\end{figure}
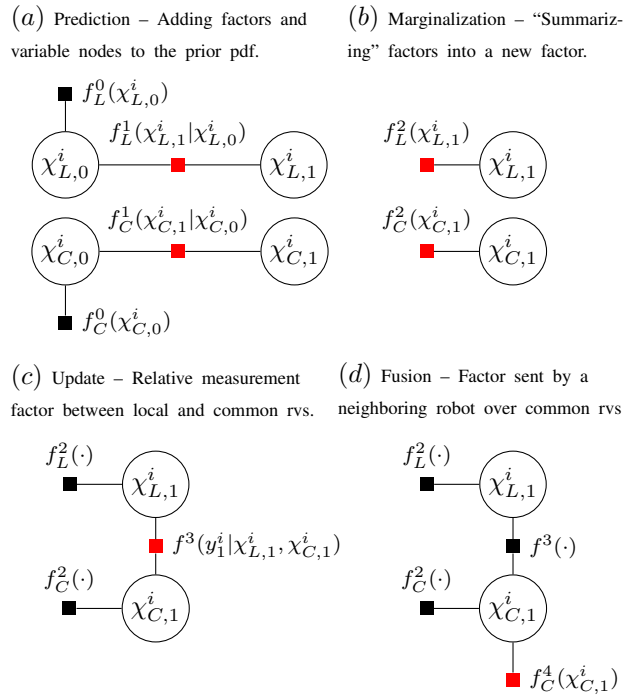

%% file: Text/4_SimulationStudy.tex
Our multi-robot multi-target experiments include both simulations and hardware experiments.
The goal is to test the robustness and applicability of FG-DDF to realistic heterogeneous data fusion challenges. 
Using Monte Carlo simulation, we test the applicability of the explicit (HS-CF) and implicit (HS-CI) heterogeneous fusion rules for a network of robots performing nonlinear dynamic multi-target tracking and self-localization. 
We then test the robustness of the algorithms to message dropouts in both simulation and linear hardware experiments, where the factor graph-based inference engine runs onboard a Clearpath Jackal UGV.


\subsection{Simulation}
We simulated two different scale problem scenarios, comprising teams of $n_r = 5$ and $n_r = 10$ robots, tracking $n_t=6$ and $n_t=12$ targets, connected in an undirected chain network ($1\leftrightarrow 2\leftrightarrow  3\leftrightarrow 4 \leftrightarrow \cdots$).
\footnote{ 
Note that a chain network is the worst case for an acyclic topology (e.g., a tree), as it takes more time for information to propagate to the ends of the network \cite{grime_data_1994}.} 
The global inference task of the team of robots is to infer the $2D$ dynamic pose, $r^i_k=[x^i_k,y^i_k,\theta^i_k]^T$ of all robots $i\in N_r$, and $2D$ dynamic target positions, $t^m_k=[x^m_k, y^m_k]^T$, of all targets $m=1,2,...,n_t$. 
The individual inference task of each robot $i$ given by $\chi^i_k=\chi_C^i\cup\chi_L^i$ is shown in Table \ref{tab:inferTasks}, with the first $5$ robot assignments being the same in both scenarios.

\begin{table}[tb]
\vspace{0.09in}
\renewcommand{\arraystretch}{1.4}
\caption{Local platform target assignments, common and local rv sets. }
    \begin{center}
    \begin{tabular}{c|c|c|c}
        Robot    & Targets&$\chi^i_C=\bigcup_{j\in N_r^i}^{}\chi^{ij}_C$&$\chi_L^i$   \\ \hline
        1 & $1, 2$& $\chi_C^{12}=t^2_k$ &$r^1_k\cup t_k^1$  \\ \hline
        2 & $2, 3$& $\chi_C^{21}=t^2_k,\ \chi_C^{23}=t^3_k$ &$r^2_k$  \\ \hline
        3 & $3, 4, 5$& $\chi_C^{32}= t^3_k,\ \chi_C^{34} = t^4_k\cup t^5_k$ &$r^3_k$  \\ \hline
        4 & $4, 5$& $\chi_C^{43} = t^4_k\cup t^5_k,\ \chi_C^{45} = t^5_k$ &$r^4_k$  \\ \hline
        5 & $5, 6$& $\chi_C^{54} = t^5_k,\ \chi_C^{56} = t^6_k$ &$r^5_k$  \\ \hline
        6 & $6, 7, 8$& $ \chi_C^{65} = t^6_k,\ \chi_C^{67} = t^7_k\cup t^8_k$ &$r^6_k$  \\ \hline
        7 & $7, 8, 9$& $\chi_C^{76} = t^7_k\cup t^8_k,\ \chi_C^{78} =  t^9_k$ &$r^7_k$  \\ \hline
        8 & $9, 10$ &$\chi_C^{87} =  t^9_k,\ \chi_C^{89} = t^{10}_k$ &$r^8_k$  \\ \hline
        9 & $10, 11$& $\chi_C^{98} = t^{10}_k,\ \chi_C^{910} = t^{11}_k$ &$r^9_k$  \\ \hline
        10 & $11, 12$& $\chi_C^{109} = t^{11}_k$ &$r^{10}_k\cup t_k^{12}$  \\ \hline
    \end{tabular}
    \end{center}
    \label{tab:inferTasks}
    \vspace{-0.2in}
\end{table}

It can be seen that at maximum, a robot estimates 9 states (e.g., robot 3) and communicates 4 (e.g., robots 3-4). 
For the larger scenario, compared to homogeneous DDF over the global $54$ states vector, this translates to $99\%$ communication and computation reduction \cite{dagan_exact_2023}.   
For further discussion and analysis of the computation and communication advantages of heterogeneous fusion compared to homogeneous fusion see \cite{dagan_exact_2023}.

\begin{table}[tb]
\renewcommand{\arraystretch}{1.2}
\caption{Mean RMSE tracking error and variance [m].  }
    \begin{center}
    \begin{tabular}{c|c|c|c}
       Robot    & Centralized & HS-CF  & HS-CI   \\ \hline
        1 & $0.26\pm0.16$ & \color{Green}{$0.29\pm0.38$} & $0.45\pm0.63$    \\ \hline
        2 & $0.23\pm0.14$ & \color{Green}{$0.38\pm0.52 $} & $0.54\pm0.74$      \\ \hline
        3 & $0.24\pm0.12$&\color{Green}{$0.32\pm0.39$}& $0.65\pm0.73$  \\ \hline
        4 &$0.31\pm0.23$ &\color{Green}{$0.47\pm0.55$}& $0.61\pm0.67$    \\ \hline
        5 &$0.55\pm0.21$ &$1.69\pm1.83$& \color{Green}{$1.63\pm1.47$}  \\ \hline
    \end{tabular}
    \end{center}
    \label{tab:smallResults}
    \vspace{-0.2in}
\end{table} 


At every time step, robots: (i) take local sensor bearing and range measurements with respect to maximum 4 known landmarks (to localize themselves) and with respect to their perspective targets; (ii) communicate factors via heterogeneous FG-DDF over subsets of common target variables $\chi^{ij}_C$ with their neighbors (see Table \ref{tab:inferTasks}). 
Note that robots do not take relative measurements of each other (as done in to cooperative localization), but their own position estimates will nevertheless get indirectly updated due to dependencies on common target positions.

The robots follow nonlinear Dubin's cars dynamics, 
\begin{equation}
    \begin{split}
        &\dot{x}^i=v^i\cos\theta^i+\omega^i_x,\\
        &\dot{y}^i=v^i\sin\theta^i+\omega^i_y,\\
        &\dot{\theta}^i = \frac{v}{L}\tan\phi^i+\omega^i_\theta,
    \end{split}
\end{equation}
where $v^i$ and $\phi^i$, and $\omega^i=[\omega^i_x, \omega^i_y, \omega^i_\theta]^T$ are the time-dependent linear velocity ($m/s$), steering angle (rad), and zero mean additive white Gaussian noise (AWGN) of robot $i$, respectively. 
$L$ is the front-rear wheel distance (taken to be $0.6m$ in the simulations).
Target $m$'s linear dynamics are modeled with an assumed known motion control law,
\begin{equation}
    \begin{split}
        &x^m_{k+1}=x^m_{k+1}+u_{x,k}^m+\omega^m_x\\
        &y^m_{k+1}=y^m_{k+1}+u_{y,k}^m+\omega^m_y,
    \end{split}
\end{equation}
where $u^m_k=[u_{x,k}^m, u_{y,k}^m]^T$ is the motion control input, and $\omega^m=[\omega^m_x, \omega^m_y]^T$ is again zero mean AWGN.
All robots were initially randomly positioned in $20m\times20m$ square and then normally sampled in each simulation with $\sigma^2=25m^2$, and target positions were randomly sampled from a normal distribution with $\sigma^2=20m^2$. 
Known landmarks were positioned in a $200m\times200m$ square. 

\subsubsection*{Explicit vs. Implicit Data Tracking} 
The first set of simulations tests the heterogeneous FG-DDF framework non-linear dynamics and measurement models. 
We performed 50 MC simulations using the HS-CF and HS-CI to explicitly and implicitly account for common data dependencies, respectively.
These dependencies can arise due to both robots using the same target dynamic model, and from previous communication episodes. 
As they are non-linearly propagated in time and `rolled up' into the current estimate upon filtering, it becomes harder to correctly remove common information during fusion according to (\ref{eq:Heterogeneous_fusion}).   
The reference for comparison is a centralized estimator marginal estimate of each robot's tracked targets, taken from the global estimate over the full $27$--state vector. 
Table \ref{tab:smallResults} summarizes the mean RMSE and variance of each robot's tracking error over its respective tracked targets.\footnote{While the  NEES chi-square consistency test \cite{rong_practical_2001}, \cite{bar-shalom_linear_2001} is a more indicative test for consistency, here we show the mean squared error, as it is visually clearer and simpler. Nevertheless, we confirmed consistency via the NEES test with results being consistent $80\%-99\%$ of the time, depending on the robot.}
As seen from the table, all robots yield a good estimate, compared to the centralized estimator, especially when considering the fact that it only requires $1\%$ of the computation costs for each robot. 
An interesting effect can be seen when comparing robot $5$ to the rest of the robots -- for robots $1-4$ the HS-CF yields smaller error and variance better, with a smaller RMSE and variance compared to the HS-CI, whereas on the other hand, the results are opposite for robot 5, with the HS-CI giving a better estimate.
This is a surprising result since recall that the homogeneous CI is an upper bound on all possible fusion covariances when considering all possible dependencies between two estimates (see \cite{julier_non-divergent_1997}) -- the HS-CI then should be an upper bound on the HS-CF results. 
One possible conclusion is that the HS-CF removes less `common data' than it should. 
As this is beyond the scope of this paper, we leave this point to future research.

\begin{figure*}[tb]
    \vspace{0.1in}
    \centering
    \includegraphics[width=0.99\textwidth]{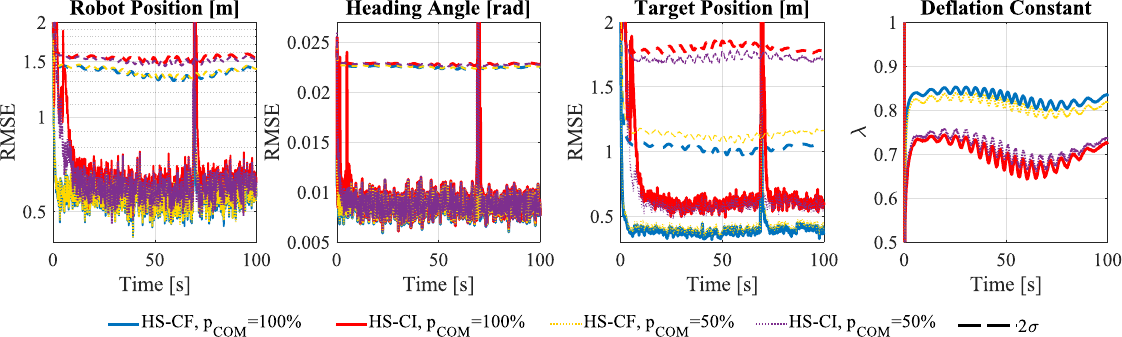}
    \caption{Simulation RMSE and $2\sigma$ confidence bounds of robot 6 for self-position, heading angle and tracking of targets 6--8, comparing: HS-CF and HS-CI fusion rules under perfect communication ($100\%$) and message dropouts -- $50\%$  communication success rate. The right figure shows the deflation constant value to guarantee conservative filtering \cite{dagan_conservative_2022}.  }
    \label{fig:nonlinearSimResults}
    \vspace{-0.2in}
\end{figure*}
\subsubsection*{Robustness}
The second set of simulations scales the FG-DDF framework to a larger scenario and tests it for imperfect communication. 
Figure \ref{fig:nonlinearSimResults} presents robot $6$'s simulation results of the 10-robot, 12-target scenario, where each robot has $100\%$ (full lines) or $50\%$ (dotted lines) probability of actually receiving messages. 
Robots do not know whether the messages they sent are received by their neighbors. 

We can see that similar to the smaller simulation scenario, the HS-CF provides better performance for both the robot's ego position estimate and the target tracking error.  
When comparing perfect communication ($100\%$) to the $50\%$ dropout rate we can see that the HS-CI fusion rule is relatively indifferent to this fact, as dependency information is not explicitly tracked. 
On the other hand, with the HS-CF, a dropped message means that the channel filters at both robots now hold a different estimate of the `common data': the sending robot CF accounts for the data sent as common, while the other robot's CF did not receive it. 
Another interesting observation can be seen from the target position graph, where the HS-CI with $50\%$ communication success rate (dotted, purple) yields better RMSE and $2\sigma$ than the perfect communication case (full, red). 
This can be explained by the lower deflation constant value for the perfect communication case, shown in the right graph, which indicates a trade-off between gaining new data from a neighboring robot vs. deflation to guarantee conservative filtering \cite{dagan_conservative_2022}.



%% file: Text/5_HardwareExperiments.tex

\begin{figure*}[htb!]
    \vspace{0.05in}
    \centering
    \includegraphics[width=0.99\textwidth]{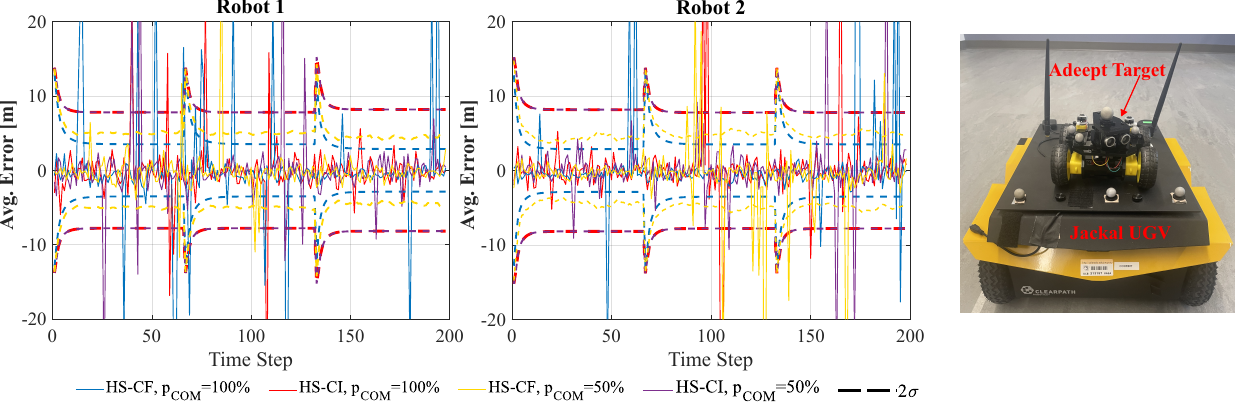}
    \caption{Average tracking error and $2\sigma$ of targets positions from hardware experiments. On the right is a picture of a tracker robot -- Clearpath Jackal UGV, and a target -- Adeept (AWR-A).}
    \label{fig:experimentResults}
    \vspace{-0.2in}
\end{figure*}

To evaluate the robustness of the FG-DDF framework in hardware implementation with respect to message dropouts and measurement outliers, we deploy it on two Clearpath Jackal UGVs, shown in Fig. \ref{fig:experimentResults}. 
The inference task of each robot is to estimate the $2D$ position and velocity of 3 out of 5 assigned targets $t^m_k=[x^m_k,\dot{x}^m_k,y^m_k,\dot{y}^m_k]^T$ ($m=1,2,..,5$), and its own constant (but unknown) robot-to-target relative position measurement bias $s^i=[b^i_{x},b^i_{y}]^T$, similar to \cite{noack_treatment_2015}. 

The Jackals robots are equipped with a 2-core Intel Celeron G1840 CPU with 4GB of RAM and 128GB of disk drive storage and a 2-core Intel i7-7500U CPU with 32GB of RAM and 512GB of disk drive storage, respectively.
Each robot runs the FG-DDF onboard as the inference and fusion engines, where ROS (version 1) is used for message passing between the robots. 
We use 5 Adeept wheeled robots for Arduino (AWR-A) as targets (see Fig. \ref{fig:experimentResults} and accompanying video). 
The targets are programmed to move in a straight line for about 4 seconds and then turn right for half a second, but due to slipping, their turn angle varies stochastically, which results in a highly nonlinear trajectory. 

In our experiments, as in many target tracking problems, the targets' dynamics are modeled using a linear `nearly constant velocity' motion model \cite{bar-shalom_linear_2001}. 
The linear relative target and landmark position measurements are gathered using Vicon motion-capture cameras, corrupted by zero mean Gaussian noise, and are modeled as,
\begin{equation}
    \begin{split}
        &y^{i,t}_{k} = t^m_k+s^i+v^{i,1}_k, \ \ v^{i,1}_k \sim \mathcal{N}(0,R^{i}),  \\
        &m^i_{k} = s^i+v^{i,2}_k, \ \ v^{i,2}_k \sim \mathcal{N}(0,R^{i}).
    \end{split}
    \label{eq:meas_model}
\end{equation}
Here the zero mean Gaussian noise for robots $1$ and $2$ is characterised by the covariance matrices $R^{1}=diag([1,10])$ and $R^{2}=diag([3,3])$, respectively.

Experiments results are shown in Fig. \ref{fig:experimentResults}. 
In all experiments, robot 1 estimates its own bias states and targets 1-3 position and velocity state, similarly, robot 2 estimates its bias and targets 3-5, the robots then have one target in common (target 3).
We performed experiments using the HS-CI and HS-CF fusion rules with different communication success probabilities and compared RMSE and $2\sigma$ bounds across each robot's 14 states, based on truth values from the Vicon system. 
Figure \ref{fig:experimentResults} compares the average tracking error of the target's east and north positions of the HS-CF and HS-CI algorithms for perfect ($100\%$) and imperfect ($50\%$) communication success rates.  

The two robots perform well, where the error spike is attributed to outliers from the Vicon measurements, which occur when targets and robots pass too close to each other. 
The $2\sigma$ lines show similar behavior to the one observed in simulations, where: 
(i) the HS-CF yields a more confident estimate than the HS-CI; 
(ii) for the HS-CF the $50\%$ communication rate yields a worse estimate; 
(iii) for the HS-CI, message dropouts have indistinguishable effects on the estimates in this scenario, which we attribute to the different measurement noise covariances between the robots, as this should cause the weight $\omega$ (\ref{eq:marginalCI}) to approach 0, i.e. almost ignoring robot $1's$ estimate. 
These experiments demonstrate the robustness of the FG-DDF to real-world effects, such as message dropout and measurement outliers, and achieve good tracking performance, despite highly nonlinear target behavior.

%% file: Text/6_Conclusions.tex
As the size of robot teams and the variety of tasks they perform increase, heterogeneous fusion becomes a core problem that must be addressed to ensure correct and scalable multi-robot information sharing for collaboration. 
Heterogeneous fusion allows robots to share only `relevant' parts of their local pdfs, dramatically 
reducing communication and computation requirements for multi-robot teams, as described in the 10-robot 12-target simulation example.  
In this paper, we test two heterogeneous fusion rules within the FG-DDF framework, namely the HS-CF, and the newly developed HS-CI, in realistic scenarios involving non-linear dynamic and measurement models, significant message dropouts, and measurement outliers. 
While under these conditions there are no formal guarantees for fusion to work (since common data dependencies are ill-defined), we show that consistent estimates can nevertheless still be produced for challenging problems like multi-target tracking with self-localization. 
From a theoretical point of view, simulation results 
suggest that the HS-CF might not always remove common data dependencies correctly for all robots in non-linear problems, and leaves an open point for future research.